\documentclass[11pt]{article}

\usepackage[preprint]{acl}

\usepackage{times}
\usepackage{latexsym}

\usepackage{booktabs}
\usepackage{soul}

\usepackage[T1]{fontenc}

\usepackage[utf8]{inputenc}

\usepackage{microtype}

\usepackage{inconsolata}

\usepackage{graphicx}

\usepackage{hyperref}
\usepackage[utf8]{inputenc}
\usepackage{paralist} 
\usepackage{diagbox} 
\usepackage{colortbl}
\usepackage{amsmath}
\usepackage{multirow}

\title{SenWiCh: Sense-Annotation of Low-Resource Languages for WiC using Hybrid Methods}

\author{
 \textbf{Roksana Goworek\textsuperscript{1,2}},
 \textbf{Harpal Karlcut\textsuperscript{1}},
 \textbf{Hamza Shezad\textsuperscript{1}},
 \textbf{Nijaguna Darshana\textsuperscript{1}},
\\
 \textbf{Abhishek Mane\textsuperscript{1}},
 \textbf{Syam Bondada\textsuperscript{1}},
 \textbf{Raghav Sikka\textsuperscript{1}},
 \textbf{Ulvi Mammadov \textsuperscript{1}},
\\
 \textbf{Rauf Allahverdiyev\textsuperscript{1}},
 \textbf{Sriram Purighella\textsuperscript{1}},
 \textbf{Paridhi Gupta\textsuperscript{1}},
 \textbf{Muhinyia Ndegwa\textsuperscript{1}},
\\
 \textbf{Bao Khanh Tran\textsuperscript{1}},
 \textbf{Haim Dubossarsky\textsuperscript{1,2,3}}
\\
\\
 \textsuperscript{1}Queen Mary University of London,
 \textsuperscript{2}The Alan Turing Institute,
 \textsuperscript{3}University of Cambridge,
\\
 }

\begin{document}

\maketitle
\begin{abstract}

This paper addresses the critical need for high-quality evaluation datasets in low-resource languages to advance cross-lingual transfer. While cross-lingual transfer offers a key strategy for leveraging multilingual pretraining to expand language technologies to understudied and typologically diverse languages, its effectiveness is dependent on quality and suitable benchmarks. We release new sense-annotated datasets of sentences containing polysemous words, spanning ten low-resource languages across diverse language families and scripts. To facilitate dataset creation, the paper presents a demonstrably beneficial semi-automatic annotation method. The utility of the datasets is demonstrated through Word-in-Context (WiC) formatted experiments that evaluate transfer on these low-resource languages. Results highlight the importance of targeted dataset creation and evaluation for effective polysemy disambiguation in low-resource settings and transfer studies. The released datasets and code aim to support further research into fair, robust, and truly multilingual NLP.

\end{abstract}


\section{Introduction}
Cross-lingual transfer is a key strategy in modern NLP, particularly for low-resource languages, where training data is scarce. By leveraging multilingual pretraining, models can transfer task-specific abilities from high-resource languages to low-resource ones, expanding access to language technologies for underrepresented communities \citep{he2021towards, ponti2018adversarial, wei2021finetuned}.

Despite its promise, transfer learning is not universally effective across tasks or languages. Studies on tasks like POS tagging, NER, NLI, QA, and sentiment analysis \citep{pires2019multilingual, dolicki2021analysing, srinivasan2021predicting, lauscher2020zero, ahuja-etal-2023-mega}, as well as polysemy disambiguation \citep{raganato2020xl, dubossarsky2024strengthening}, show that cross-lingual transfer can be inconsistent and, in some cases, fail entirely. This is also true for generative models \citep{robinson2023chatgpt, shaham2024multilingual, chirkova2024zero}, with particularly poor performance in low-resource languages, highlighting the need for more robust and language-inclusive transfer.

A main obstacle for transfer is the lack of high-quality datasets in low-resource and typologically diverse languages. Without these benchmarks, assessing transfer performance, let alone training models on target languages, remains a formidable challenge. This lacking is largely due to the scarcity of linguistic resources in low-resource languages. For instance, Wiktionary contains over a million entries for German, English, French, Chinese, and Russian, but fewer than 100,000 for Punjabi and Marathi \citep{WikipediaListofWikipedias}.

This lack of resources underscores the urgent need for dedicated datasets to evaluate and refine transfer techniques for underrepresented languages, which this work addresses by developing a semi-automatic method for sense annotation in polysemy and generating resources in ten languages. 

We focus on the task of polysemy disambiguation, as it particularly challenges cross-lingual transfer by revealing structural and semantic differences between languages.
While some NLP tasks, like sentiment analysis, rely on meaning preservation across languages, where direct translation can maintain performance, polysemy is highly language-specific \citep{rzymski2020database}, making it a rigorous test of a model’s ability to generalize across languages. For example, the English word "movement" refers to both physical motion and a political or social movement. However, its Polish translation, "ruch", also encompasses these two meanings, but additionally means "traffic", a sense not covered by the English word. Conversely, "movement" in English can also refer to a section of a musical composition.


Polysemy disambiguation has long been considered a hallmark of human cognition and a central challenge in NLP \citep{navigli2009word, bevilacqua2021recent}. A model that can accurately distinguish between different senses of a word must capture linguistic subtleties, metaphorical meanings, and even emerging word usages, much like human speakers. Thus, success in cross-lingual polysemy disambiguation would suggest a model's ability to generalize deep semantic understanding, beyond surface-level patterns in a single language.
While many high-resource languages already benefit from sense-annotated datasets (see \S\ref{sec:related work}), low-resource languages remain largely unrepresented in this area. Existing contextualized models can process polysemous words within downstream tasks \citep{loureiro2021analysis, ushio2021bert}, but sense disambiguation remains a major challenge across dozens of languages \citep{pilehvar2018wic, raganato2020xl, martelli2021semeval, liu2021am2ico}.

Beyond NLP, polysemy also presents difficulties in multimodal models, such as object detection systems, where the same word can refer to multiple visual categories \citep{calabrese2020fatality}. This suggests that solving polysemy is not just beneficial for language tasks but has broader implications for AI reasoning and multimodal understanding.



\paragraph{Our Contributions}
Despite extensive work on polysemy disambiguation in high-resource languages, datasets for low-resource languages remain scarce. We address this gap with the following contributions:

\begin{itemize}
    \item \textbf{Sense-annotated datasets:} We release both WSD-style sense-annotated corpora and WiC-style evaluation datasets for ten low-resource languages.\footnote{available at \href{https://doi.org/10.5281/zenodo.15493005}{DOI: 10.5281/zenodo.15493005}} The WiC format supports direct comparison with existing experiments in other languages, enabling strong cross-lingual baselines.
    
    \item \textbf{Annotation tool:} To facilitate further resource development, we release a hybrid semi-automated annotation tool.\footnote{available at \href{https://github.com/roksanagow/projecting_sentences}{github.com/roksanagow/projecting\_sentences}}
\end{itemize}

Together, these contributions represent a crucial step toward advancing fair, robust, and truly multilingual NLP by enabling evaluation and development in languages that have been largely neglected.

\section{Related Work}
\label{sec:related work}
\subsection{Transfer Studies}
Zero-shot cross-lingual transfer has been widely studied, with mixed findings on its effectiveness, particularly in polysemy disambiguation. While some studies highlight transfer potential across languages, others expose significant limitations, especially in tasks that depend on fine-grained semantic distinctions.

\citet{lauscher2020zero} examined zero-shot transfer performance across 17 languages and five NLP tasks (excluding polysemy), evaluating XLM-R \cite{conneau-etal-2020-unsupervised} and mBERT \citep{hf-mbert}. They found that zero-shot performance drops significantly compared to full-shot settings and that transfer success correlates with factors like pretraining corpus size and linguistic similarity. These findings suggest that cross-lingual transfer is far from universal and is highly dependent on language resources and pretraining coverage.

Focusing specifically on polysemy disambiguation, \citet{raganato2020xl} conducted the first large-scale cross-lingual transfer study for this task, training a model on English and evaluating on 12 other languages. While they observed some zero-shot transferability, models trained on English underperformed models trained on the target language by 10-20\% when tested on German, French, and Italian, indicating that polysemy disambiguation remains language-sensitive and benefits from in-language supervision.

In contrast, \citet{dubossarsky2024strengthening} challenged the feasibility of cross-lingual transfer for polysemy disambiguation altogether. Studying English and Hindi, they found a complete lack of zero-shot transfer, suggesting that word sense distinctions may be too language-specific for direct transfer without explicit in-language supervision.

These conflicting results emphasize the need for more comprehensive transfer studies in polysemy disambiguation, particularly in low-resource languages where transfer learning is often the only viable approach due to the lack of labeled data. However, without high-quality evaluation datasets in these languages, assessing and improving transfer learning for polysemy remains an open challenge.

\subsection{Polysemy Disambiguation}  


Word Sense Disambiguation (WSD) datasets are sense-annotated corpora consisting of sentences containing polysemous words, labeled according to their contextual meanings. WSD is inherently complex, as words vary in the number of possible senses, and the list of words differs across languages. 
To address this, \citet{pilehvar2018wic} introduced the Word in Context (WiC) formulation, which reformulated the original WSD problem, which was a multi-class classification task, into a binary classification one. Instead of assigning specific sense labels, WiC pairs two sentences containing the same word and labels them 1 (same) or 0 (different). For example:

\begin{quote}
A \textbf{bat} flew out of the cave as the sun set. \newline
He swung the \textbf{bat} with all his strength.
\end{quote}
\label{sec:wic example}

This approach enables models to be trained directly on polysemy disambiguation by adjusting embeddings so that words with the same sense cluster together, while those with different senses are pushed apart in the resulting embedding space.  





\subsection{Existing Datasets}
\subsubsection{WSD Datasets}

Word Sense Disambiguation (WSD) research has been supported by several key sense-annotated corpora and lexical resources:
\newline
\textbf{SemCor} \citep{miller1993semantic} is a foundational English corpus containing over 226,000 sense annotations across 352 documents. 
\newline
\textbf{OntoNotes} \citep{hovy2006ontonotes} offers a multi-genre corpus with extensive annotations, including word senses linked to a refined sense inventory for English, Chinese and Arabic.
\newline
\textbf{Senseval/SemEval Datasets} have been instrumental in standardizing WSD evaluation. Notably, \textbf{Senseval-2} \citep{edmonds2001senseval} and \textbf{SemEval-2007 Task 17} \citep{pradhan-etal-2007-semeval} provided all-words WSD tasks, challenging systems to disambiguate every content word in given texts. These competitions have included data in multiple languages, such as English, Chinese, Basque, and others \citep{navigli-etal-2013-semeval}.
\newline
\textbf{CoarseWSD-20} \citep{loureiro2021analysis} is a coarse-grained sense disambiguation dataset derived from Wikipedia, focusing on 20 ambiguous nouns, each with 2 to 5 senses, all in English. 
\newline
\textbf{FEWS (Few-shot Examples of Word Senses)} \citep{blevins2021fews} addresses the challenge of disambiguating rare senses. Automatically extracted from Wiktionary, FEWS provides a large training set covering numerous senses and an evaluation set with few- and zero-shot examples, facilitating research in low-shot WSD scenarios in English.
\newline
\textbf{WordNet} \citep{miller1995wordnet} serves as a comprehensive lexical database grouping words into synsets representing distinct concepts. Each synset is interconnected through various semantic relations, offering a structured sense inventory integral to WSD tasks. It primarily focuses on English, but various projects have extended it to other languages.
\newline
\textbf{BabelNet} \citep{navigli2012babelnet} extends WordNet by integrating it with Wikipedia and other resources, forming a multilingual semantic network. As of version 5.3 (December 2023), BabelNet covers 600 languages, containing almost 23 million synsets and around 1.7 billion word senses \citep{babelnet2023}. This expansive resource connects concepts across languages, supporting cross-lingual WSD and enriching the sense inventory beyond monolingual constraints.

\subsubsection{WiC Datasets}

The Word-in-Context (WiC) framework has been instrumental in evaluating context-sensitive word embeddings through binary classification tasks. Several notable datasets have been developed within this framework:

\textbf{WiC} \citep{pilehvar2019wic} is the pioneering English dataset that introduced the WiC framework. It consists of sentence pairs where a target word appears in both contexts, and the task is to determine whether the word carries the same meaning in both sentences. This dataset has set the standard for subsequent WiC-based evaluations.

\textbf{XL-WiC} \citep{raganato2020xl} extends the WiC framework to a multilingual setting, encompassing 12 languages: Bulgarian, Danish, German, Estonian, Farsi, French, Croatian, Italian, Japanese, Korean, Dutch, and Chinese. This expansion facilitates cross-lingual evaluation of semantic contextualization and enables research into zero-shot transfer capabilities of multilingual models.

\textbf{MCL-WiC} \citep{martelli2021semeval} offers datasets in English, Arabic, French, Russian, and Chinese. These were constructed by annotating sentences from native corpora, including BabelNet \citep{navigli2012babelnet}, the United Nations Parallel Corpus \citep{ziemski-etal-2016-united}, and Wikipedia. The dataset achieved inter-annotator agreements of 0.95 and 0.9 for English and Russian, respectively, indicating high annotation quality.

\textbf{AM\textsuperscript{2}iCo} \citep{liu2021am2ico} presents a multilingual dataset pairing English with 14 target languages. Compiled from Wikipedia dumps of each language, it selects words with at least two distinct pages, indicating ambiguity in both the target language and English. The dataset reports an overall human accuracy of 90.6\% and an inter-annotator agreement of 88.4\%.

\textbf{WiC-TSV} \citep{breit2021wic} introduces a multi-domain evaluation benchmark for WiC, independent of external sense inventories, but only in English. Covering various domains, WiC-TSV provides flexibility for evaluating diverse models and systems both within and across domains.

Despite these advancements, there remains a significant gap in resources for low-resource languages. Our dataset aims to address this deficiency by providing sense-annotated data in both WSD and WiC formats for underrepresented languages, thereby facilitating research in polysemy disambiguation and cross-lingual transfer across a broader spectrum of linguistic contexts.

\section{Methods}

\subsection{Dataset Curation}

We follow the below method for the curation of sense-annotated datasets, adjusted for language-specific considerations. These are detailed in section \S\ref{sec:language specific}, along with the resources used for the curation of the dataset in each language.

\noindent\paragraph{1. Identification of Polysemous Words} Publicly available dictionaries (online and offline) were surveyed. By searching for words with more than a single dictionary entry, lists of hundreds of candidate polysemous words were compiled. Where available, lists of polysemous words were added.

\noindent \paragraph{2. Corpus Selection and Sentence Sampling} Native corpora of sufficient size were chosen to ensure diverse contextual representation of target words. Candidate polysemous words were filtered based on corpus frequency, removing low-frequency terms, and manually reviewed for sense granularity. From these corpora, large samples of sentences (typically 100-1000 per word) were randomly extracted for further analysis.

\noindent \paragraph{3. Embedding-Based Analysis} 
Word embeddings were generated for target words in the sampled sentences, and dimensionality reduction methods and clustering techniques were applied to these to create interactive 2D visualizations (see \S\ref{subsect:semiAutomaticAnnotation}).

\begin{figure}
    \centering
    \includegraphics[width=1\linewidth]{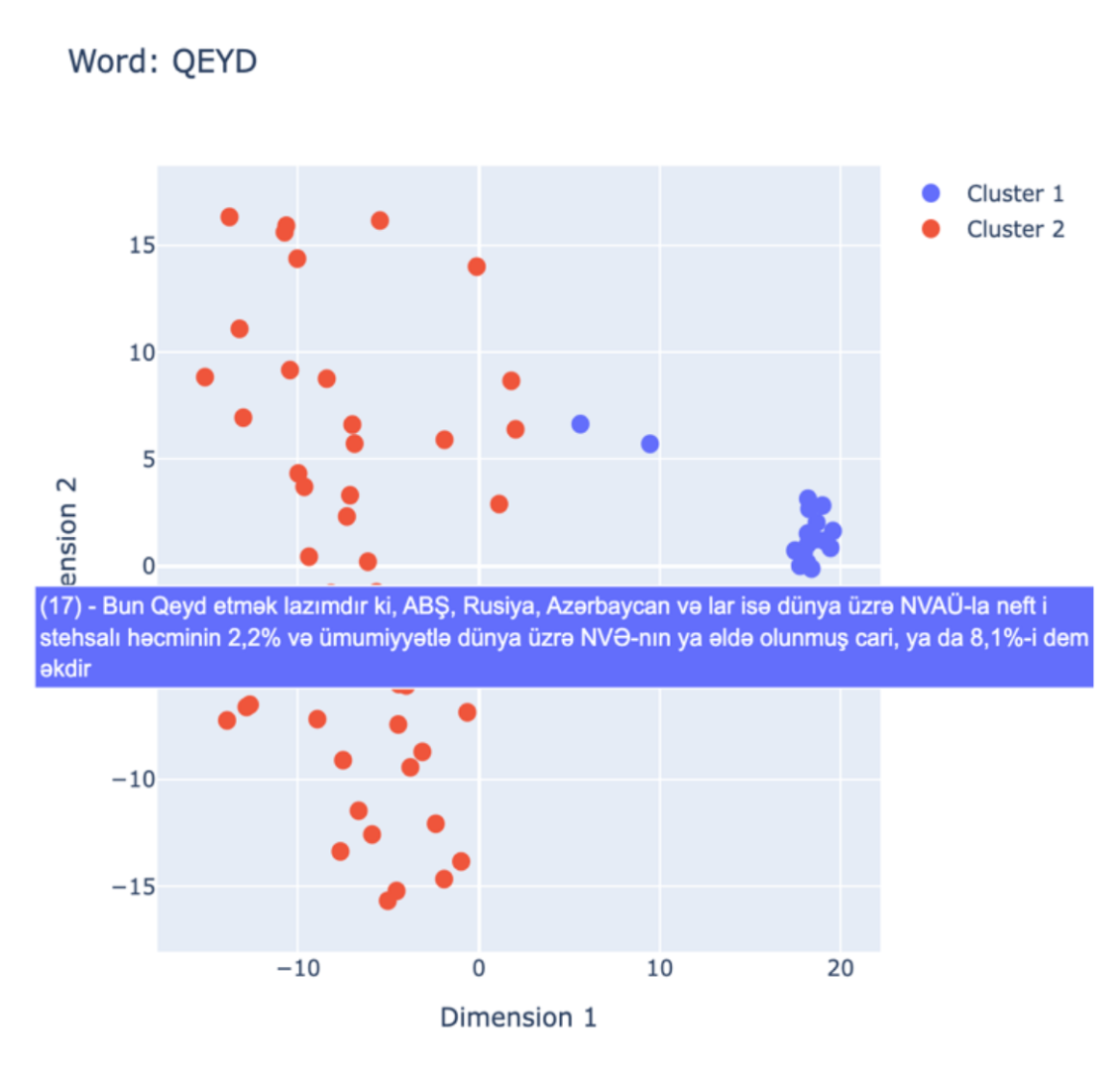}
    \caption{Example of interactive embedding-based sentence selection for the Azerbaijani word `qeyd'.}
    \label{fig:2d-viz}
\end{figure}

\noindent \paragraph{4. Manual Annotation of Sentences:} In the 2D visualization, presented in Figure \ref{fig:2d-viz}, annotators could hover over points representing sentences and click to assign them to different sense groups, for one word at a time. Sentences were selected based on their distribution in the embedding space or automatic clustering labels, with priority given to those that were more dispersed to ensure broad semantic coverage and enhance the representation of rare senses.

\subsection{Semi-Automatic Annotation Tool}
\label{subsect:semiAutomaticAnnotation}

Our annotation process is semi-automatic, using vector representations for efficient sentence selection while ensuring manual verification.

To represent sentences in a structured way, we embed usages of the target word in all candidate sentences using pretrained transformer-based models such as mBERT \cite{devlin2019bert}, XLM-R \cite{conneau-etal-2020-unsupervised}, or language-specific models. These embeddings capture contextual semantics, making them suitable for sense-based clustering. We then apply K-Means or agglomerative clustering to group sentences into distinct senses, followed by dimensionality reduction techniques (e.g., UMAP, MDS) to visualize their distribution in 2D space (see Figure \ref{fig:2d-viz}).

This visualization allowed annotators to interact with embeddings, exploring clusters and selecting diverse sentences that represent different word senses. 
This is essential for identifying sentences that correspond to rare word senses, as manually searching through randomly sampled sentences would be time-consuming and often ineffective, requiring the review of an extensive number of sentences to find relevant sentences.



\subsection{Evaluating Annotation Efficiency}
Annotating subordinate senses in polysemy is inherently time-consuming due to their rarity. Since these senses occur infrequently, manually identifying them requires scanning a large number of sentences before encountering a relevant instance. 

The exact effort depends on the prior probability of the subordinate sense: the rarer it is, the more sentences need to be reviewed. To establish these priors, we randomly sampled 100 sentences for manual inspection to determine sense distributions. We then assessed how well model-based sentence selection captures each sense by comparing the proportion of automatically selected sentences correctly assigned to a sense against the baseline probability of encountering that sense in the corpus.

Our results demonstrate that computational methods significantly reduce this burden. We evaluate their effectiveness using adjusted \textbf{Lift}, a metric from Data Mining that measures improvement over random selection:

\[
\text{Lift(sense)} = \frac{Precision(sense)}{Prior(sense)} 
\]

\noindent where \textit{Precision(sense)} is the proportion of correctly classified sentences for the sense, and \textit{Prior(sense)} their probability of occurrence in the dataset. Higher Lift values indicate a greater efficiency gain in selecting rare senses.

For example, in Kannada, identifying the word {\includegraphics[height=0.9em]{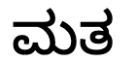}} in its subordinate ‘religion’ sense yielded a Lift of 900\%, meaning that the likelihood of finding relevant sentences increased ninefold compared to random selection. Given a prior distribution of 96:4 favoring the dominant sense, manual selection would require reviewing 25 sentences on average to find one relevant case. With automatic selection achieving 36\% precision, only three selections are needed—an 8× reduction in effort.

This efficiency boost translates directly into time and cost savings. If manual annotation takes 30 seconds per sentence, annotating 1,000 examples of a rare sense would traditionally require 8 hours of labor. With our automated method, this drops to about an hour, dramatically reducing annotation costs and making large-scale sense labeling more feasible. In Table \ref{tab:pivc_results}, we present the Lift scores for the senses of two words in each of the four languages. Additional results, covering five words for each of these languages, are provided in Table \ref{tab:pivc_more} in the Appendix, covering all words selected for this evaluation.

\begin{table}[ht]
    \centering
    \renewcommand{\arraystretch}{1.2}
    \small
    \begin{tabular}{|*{6}{c|}}
        \hline
        \multirow{2}{*}{\textbf{Lang}} & \multirow{2}{*}{\textbf{Word}} & \multicolumn{2}{c|}{\textbf{Sense Definitions}}  
        & \multicolumn{2}{c|}{\textbf{Lift (\%)}} \\ 
        \cline{3-6}
        & & 1 & 2 & 1 & 2 \\ 
        \hline
        \multirow{2}{*}{\textbf{KN}} & {\includegraphics[height=0.85em]{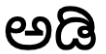}} & Foot & Under & 269 & 141 \\ \cline{2-6}
        & {\includegraphics[height=0.85em]{latex/figs/kn_w2.png}} & Opinion & Religion & 104 & 900 \\ \hline
        \multirow{2}{*}{\textbf{MR}} & {\includegraphics[height=0.85em]{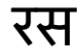}} & Juice & Interest & 128 & 188  \\ \cline{2-6}
        & {\includegraphics[height=0.85em]{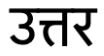}} & Answer & North & 121 & 125 \\ \hline
        \multirow{2}{*}{\textbf{PA}} & {\includegraphics[height=0.85em]{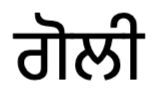}} & Bullet & Pill   & 107 & 1364 \\ 
        \cline{2-6}
        & {\includegraphics[height=0.85em]{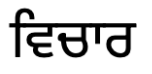}} &  Thought & Intention & 235 & 884 \\ 
        \hline
        \multirow{2}{*}{\textbf{UR}} & {\includegraphics[height=0.85em]{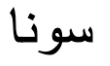}} &  Gold & Sleep & 161 & 232  \\ \cline{2-6}
        & {\includegraphics[height=0.85em]{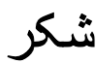}} &  Thanks & Sugar  & 106 & 1414 \\ \hline
        
    \end{tabular}%
    
    \caption{Measured improvement over random chance (Lift) in semi-automated sentence selection.}
    \label{tab:pivc_results}
\end{table}

\section{Sense-annotated Datasets}



We introduce a sense-annotated corpus of sentences containing polysemous words covering ten low resource languages that span different language families and use different scripts: Azerbaijani (Turkic), Kannada and Telugu (Dravidian), Punjabi, Marathi and Urdu (Indo-Aryan),  Polish (Slavic), Swahili (Afro-semitic), Vietnamese (Austroasiatic) and Korean (Koreanic).
Statistics for each language are presented in Table \ref{tab:dataset_statistics}.

\begin{table*}[!h]
\centering
\begin{tabular}{lrrrll}
\toprule
 Language (ISO) & Words & Sentences & Senses & Avg. Senses/Word & Avg. Sentences/Sense \\
\midrule
Azerbaijani (AZ) & 60 & 4214 & 119 & 1.98 ± 0.13 & 35.55 ± 6.09 \\
Kannada (KN) & 59 & 4446 & 127 & 2.15 ± 0.45 & 35.01 ± 14.07 \\
Korean (KO) & 28 & 1013 & 58 & 2.07 ± 0.54 & 17.81 ± 4.73 \\
Marathi (MR) & 63 & 3766 & 125 & 1.98 ± 0.13 & 30.16 ± 2.72 \\
Polish (PL) & 66 & 2877 & 158 & 2.39 ± 0.68 & 18.28 ± 5.22 \\
Punjabi (PA) & 55 & 4969 & 127 & 2.31 ± 0.54 & 39.25 ± 1.89 \\
Swahili (SW) & 22 & 1376 & 46 & 2.09 ± 0.29 & 29.91 ± 4.39 \\
Telugu (TE) & 51 & 4534 & 100 & 1.96 ± 0.28 & 45.37 ± 7.83 \\
Urdu (UR) & 39 & 2674 & 90 & 2.31 ± 0.52 & 29.72 ± 1.06 \\
Vietnamese (VI) & 11 & 1021 & 29 & 2.64 ± 0.81 & 36.14 ± 19.20 \\
\bottomrule
\end{tabular}
\caption{Statistics and ISO codes for the Multilingual WSD Sense-Annotated Dataset.}
\label{tab:dataset_statistics}
\end{table*}

\subsection{Language Specific Treatment}
\label{sec:language specific}

For each language, the dataset was compiled and annotated by native speakers with the support of computational methods described above. 

\textbf{Azerbaijani:}
Polysemous words were selected from Azerbaycan Dilinin Omonimler Lugeti \cite{hasanov2007azerbaycan}, and sentences containing selected target words were sampled from \href{https://huggingface.co/datasets/azcorpus/azcorpus_v0}{AzCorpus}, the largest open-source NLP corpus for Azerbaijani \cite{azcorpus_v0}. 
Three models were used to embed sentences: XLM-R, BERT-Turkish \cite{DBMDZ_BERT_Turkish}, and XL-LEXEME \cite{cassotti2023xl}.





\textbf{Kannada:}
Polysemous words were selected from the online Kannada dictionary \cite{venkatasubbaiah1981kannada}.  \citet{kakwani-etal-2020-indicnlpsuite} was used as a corpus, which was preprocessed to remove extraneous characters, symbols, non-linguistic patterns, excessively long or single-word sentences, and duplicate entries. Initially, sentences for five words were annotated manually. Next, Claude 3.5 Sonnet \cite{claude35sonnet} was used to pre-label sentences after demonstrating reliable performance on the manually annotated data.
The model, given Kannada and English meanings for each word, classified sentences containing the remaining target words. This streamlined human annotation, as annotators selected 30-40 sentences per sense from Claude's labels, rather than relying on clustering or embeddings for sentence selection. Finally, an independent reviewer verified all annotations.







\textbf{Korean:}
The Korean Dictionary of \citet{binjang_NIKL_korean_english_dictionary} was used to extract list of polysemous words. Two corpora were used for sampling sentences: the Korean Wikipedia Dataset \cite{lee2024wikipedia_korean_dataset} and KoWikiText \cite{kim2020kowikitext}.
A Korean contextualized model \cite{Ham2020KoSRoberta} was used to embed sentences. 




\textbf{Marathi:}
The Marathi-English Dictionary from the Digital South Asia Library (DSAL) \cite{molesworth1857marathi} was used to select polysemous words. 
For sampling sentences, three corpora were used: The Full Marathi Corpus \cite{joshi2022mahanlp}, and Marathi portions of two Indic corpora \cite{kakwani2020indicnlp, kumar2023sangraha}. 
MuRIL \cite{khanuja2021muril}, mBERT, IndicBERT \cite{kakwani2020indicnlp}, XLM-R, and XL-LEXEME were used for embedding sentences.






\textbf{Polish:}
Polysemous words were identified by reviewing native texts, verified using the Polish Online Dictionary \cite{sjp_pwn_online}, and selected if they had distinct senses.
Three corpora covering distinct domains—national corpus, news, and literature—were used to sample sentences \cite{degorski2012recznie, LeipzigPolNews2018, lebedev2023polishcorpus}. 
XL-LEXEME and a Polish BERT \cite{kleczek2020bertpolish} were used for embedding sentences.
Given Polish's high degree of inflection-where nouns, adjectives, and verbs vary by case, number, gender, and aspect across seven grammatical cases-all corpora were lemmatized to find sentences with target words in their base form for sentence selection and then restored to their original form for manual annotation. 

\textbf{Punjabi:} Only text in Gurmukhi script was considered. Polysemous words were selected from previous work on WSD in Punjabi \cite{singh2018naive, singh2019sense, singh2020word, singh2015word} as well as from dictionaries \cite{joshi2009punjabi, goswami2000punjabi, singhbrothers2006punjabi}. Sentences were sampled from Metatext \cite{conneau-etal-2020-unsupervised}, Samanantar \cite{Ramesh2022} and Sangraha \cite{Khan2024}. MuRIL, IndicBERT, mBERT, XLM-R and XL-LEXEME were used to embed the sentences.

\textbf{Swahili:}
The Swahili Dictionary \cite{kamusi2013kiswahili} was used to identify polysemous words, while the Swahili Corpus by \citet{MASUA2024110751} provided sentences. Multiple models were used for embedding (XLM-R, BERT and mBERT), but SwahBERT \cite{martin-etal-2022-swahbert} outperformed them on the initial annotated dataset and was used to aid further annotation.





\textbf{Telugu:}
Three corpora were used for selecting polysemous words, two Indic corpora \cite{kunchukuttan2020ai4bharat, kakwani2020indicnlp} and the corresponding Wikipedia Dump \cite{wikipedia2024telugu}. 
The same Indic corpus \cite{kunchukuttan2020ai4bharat} was used for sentence selection, along with the Leipzig Telugu Corpus \cite{tel_community_2017}.
For embeddings, TeluguBERT \cite{joshi2022l3cubehindbert} and MuRIL were compared, with the former outperforming. 




\textbf{Urdu:}
Two word sense-annotated corpora \cite{saeed2019wsd, saeed2019sense}, the Urdu Wiktextract \cite{ylonen2022wiktextract}, and a publicly available vocabulary book \cite{bruce2021urdu} were used to select polysemous words. The Urdu Monolingual Corpus (UrMono) \cite{jawaid2014tagged} was used to sample sentences. For embedding, mBERT, XLM-R, MuRIL, and XL-LEXEME were tested with the latter outperforming the rest. Given Urdu’s complex inflectional morphology and honorific system, a list of up to six inflected forms was generated for each noun, considering variations in number, gender, and case to ensure a diverse sentence selection.

\textbf{Vietnamese:}
Polysemous words were selected from the Tuttle Concise Vietnamese Dictionary \cite{giuong2014tuttle}, while sentences containing target words were sampled from the English-Vietnamese Parallel Corpus (EVBCorpus) \cite{ngo2013evbcorpus}.
For embedding, XL-LEXEME, XLM-R, mBERT, as well as two Vietnamese-specific models, PhoBERT \cite{nguyen-tuan-nguyen-2020-phobert} and ELECTRA \cite{nguyen2025ner_vietnamese_electra} were evaluated. As with other languages, PhoBERT emerged as the best model, highlighting the need for language-specific methods and resources.










\subsection{WiC Pairing}
\label{sec:wic algorithm}
For model training we convert the sense-annotated data in each language to the WiC format (see \S\ref{sec:wic example}).

To guarantee that the train-dev-test splits contain well-representative samples of words and sentences, and ensure sentences appear only in a single split, we use the following steps to convert sense-annotated sentences to WiC sentence pairs:

\noindent \paragraph{1. Word Splitting} 70\% of the words are randomly allocated to the training set, while 15\% each are allocated to validation and test sets.
   
\noindent \paragraph{2. Sentence Redistribution} 30\% of words from the training set are randomly selected to appear in all three splits (each sentence appearing only in one of the splits).
For these words, 25\% of their sentences are reallocated to the validation and test sets, ensuring: (1) Equal distribution between sets; (2) No sentence overlap across splits; and (3) The distribution of senses remains unchanged.

\noindent \paragraph{3. Pairing Sentences into WiC Pairs} Within each split, each sentence is paired with up to 16 different sentences, ensuring a balanced mix of same-sense and different-sense pairs.
 
\noindent
The amounts were selected to approximate a 70-15-15 dataset split. 
This approach ensures a representative, well-distributed, and balanced dataset for WiC training and testing, although it's important to note that different random seeds for sampling can result in different results, especially for smaller datasets.
Descriptive statistics of the resulting WiC datasets can be found in Table \ref{tab:wic statistics} in the Appendix. All sets are approximately balanced, setting chance performance close to 50\%.



\begin{table*}[]
\begin{tabular}{|l|cccccccccc||c|}
\hline
    \diagbox{Condition}{Test Lang} & AZ & KN & KO & MR & PL & PA & SW & TE & UR & VI & Avg. \\ \hline \hline
    Full-shot & 65.9 & 65.9 & 56.4 & 83.2 & 72.3 & 65.9 & 59.5 & 63.8 & 68.8 & 57.2 & 65.9\\ \hline 
    Zero-shot & 66.3 & \textbf{72.3} & 64.2 & 82.2 & \textbf{79.1} & 70.5 & 68.6 & 62.4 & \textbf{74.0} & \textbf{70.6} & 71.0 \\ \hline 
    Mixed & \textbf{71.9} & 71.0 & \textbf{66.5} & \textbf{88.1} & 65.4 & \textbf{81.6} & \textbf{76.9} & \textbf{65.4} & 64.8 & 68.4 & \textbf{72.0 }\\ \hline 
    
    

\end{tabular}
\caption{Accuracies of XLM-R models evaluated on the test sets of our WiC datasets. Full-shot refers to models trained exclusively on the target language's training data. Zero-shot results correspond to XLM-R trained only on English WiC data. Mixed models are first trained on English, then fine-tuned on the target language.} 
\label{tab:results}
\end{table*}
\section{Experiments}
To assess the quality of the datasets we created, and to demonstrate the need for proper evaluation in low-resource languages, we tested transfer in three transfer conditions, full-shot, zero-shot and mixed. The \textbf{full-shot} condition is mainly a sanity-check, and serves to evaluate the quality of the training set, as it does not test for transfer. In \textbf{zero-shot}, a model is fine-tuned on English (combined training data taken from the MCL \cite{martelli2021semeval} and XL \cite{raganato2020xl} datasets, totaling 13.4k sentence pairs) and evaluated on each of our ten languages, which it was not fine-tuned on. In the \textbf{mixed} condition, a model is first fine-tuned on English, and then on the target language training data, evaluating on the target language. This allows us to investigate whether leveraging large amounts of data in a high-resource language can enhance full-shot performance on low-resource corpora.




We use XLM-RoBERTa \citep{conneau-etal-2020-unsupervised} due to its strong multilingual capabilities. The model is pretrained on 100 languages, including all those in our novel datasets. It has proven highly effective in embedding both high- and low-resource languages and is widely studied in cross-lingual transfer research \citep{philippy-etal-2023-towards}, particularly in the context of polysemy disambiguation \citep{raganato2020xl, dubossarsky2024strengthening, cassotti2023xl}.



For model fine-tuning, we follow \citet{cassotti2023xl} and use a bi-encoder architecture that independently processes the two sentences containing the polysemous target word using a Siamese network to generate two distinct vector representations (embeddings).
The model outputs the cosine distance between the output embeddings of the two inputs, and, to collapse this to a binary label, a threshold is applied to decide if the words are classified as having the same sense. The model is trained to adapt embeddings and increase this distance when the target word has different meanings and decrease it when the meanings are the same in the two sentences by minimising contrastive loss.
After training, we set the threshold for each model by maximising accuracy on the corresponding validation set. 
During training, as well as inference, special tokens, \texttt{<t>} and \texttt{</t>}, are placed around the target word in each sentence to signal what word the model should focus on.




\section{Results}

\noindent \paragraph{Our semi-automatic annotation method works} The transfer results (Table \ref{tab:results}) demonstrate that we were able to produce high-quality datasets in ten low-resource languages. The low performance in Korean, Swahili, and Vietnamese is only observed in the full-shot condition. These are most likely due to their smaller training size rather than quality issues; otherwise, low performance would have been observed also in the zero-shot condition.

\noindent \paragraph{Evaluating on all target languages is essential} Transfer effects are not uniform, as seen in the zero-shot performance that varies from 62.4\% in Telugu to 82.2\% in Marathi. Interestingly, zero-shot outperforms full-shot in 8 out of 10 languages, and gets comparable accuracy in the remaining 2, likely due to the small training data size of full-shot models and strong transfer from English. These results emphasize the unpredictability of transfer from one side, but also stress the need for a comprehensive multilingual benchmark to accurately assess cross-lingual transfer and ensure models perform reliably across diverse languages. With our efficient semi-automatic annotation method, curating such datasets is also much cheaper in annotation efforts.


\noindent \paragraph{Mixed training improves transfer} For most languages, mixed-training improves upon either full-shot or zero-shot conditions. This hybrid strategy leverages large-scale training data in English with language-specific details from the target language for effective polysemy resolution. This further highlights the importance of datasets in low-resource languages, where even small amounts of labeled data can lead to marked improvements.

\section{Discussion}

In this work we present sense-annotated datasets across a diverse range of language families, providing valuable resources for linguistic and computational studies. Punjabi, Marathi, and Urdu belong to the Indo-Aryan branch, enabling research on linguistic relatedness alongside the Hindi WiC dataset \cite{dubossarsky2024strengthening}. Telugu and Kannada represent the Dravidian family, while Azerbaijani, Swahili, Vietnamese, Polish, and Korean extend coverage to additional linguistic groups.
The dataset includes Arabic-based (Punjabi, Urdu), Devanagari (Marathi), Latin-based (Azerbaijani, Polish, Swahili, Vietnamese), Hangul (Korean), and Brahmic scripts (Kannada, Telugu), facilitating research on script variation and its impact on NLP.

By encompassing a broad linguistic spectrum, our dataset supports studies on linguistic relatedness, historical evolution, and polysemy disambiguation in low-resource settings. It serves as a foundation for evaluating and improving multilingual and cross-lingual transfer, particularly in tasks requiring deep semantic understanding.

Our experiments highlight the importance of language-specific resources. The unexpected finding that zero-shot XLM-R trained only on English outperformed full-shot models trained on the target language challenges assumptions about cross-lingual transfer stability, emphasizing the need for dedicated evaluation datasets.

Manual annotation is essential yet labor-intensive, particularly for low-resource languages. We introduce an automated method to identify sentences across all word senses, even when certain senses are sparsely represented. Our quantitative results demonstrate the effectiveness of this approach in enhancing annotation efficiency and supporting sense-annotated dataset development. To encourage further research, we release our code on GitHub: \href{https://github.com/roksanagow/projecting_sentences}{github.com/roksanagow/projecting\_sentences}.

\section{Limitations}




The dataset remains relatively small, which may limit the generalizability of findings, particularly for full-shot experiments, where additional training data would likely improve performance. Additionally, data imbalance across languages makes direct comparisons challenging without subsampling, which in turn reduces overall performance. Even within a single language, the number of senses and sentences per word varies, further complicating evaluation. Moreover, each language was sourced from different corpora, leading to potential inconsistencies in text style, domain coverage, and annotation quality.

The evaluation setup also has certain constraints. Train-dev-test splits were generated randomly (according to the algorithm specified in \S\ref{sec:wic algorithm}), and the prevalence of sentences corresponding to different words across splits could impact the results. Furthermore, zero-shot evaluation was conducted only from English, leaving open questions about transfer from other high-resource languages and cross-lingual settings beyond English-centric transfer.

\section{Acknowledgments}
This work was partially funded by the research program Change is Key!, supported by Riksbankens Jubileumsfond (reference number M21-0021). The authors would like to thank Bao Linh Hoang for contributing additional expert annotations.

\appendix

\newpage
\section{Example visualization of annotated sentences}
\begin{figure}[h]
    \centering
    \includegraphics[width=0.8\linewidth]{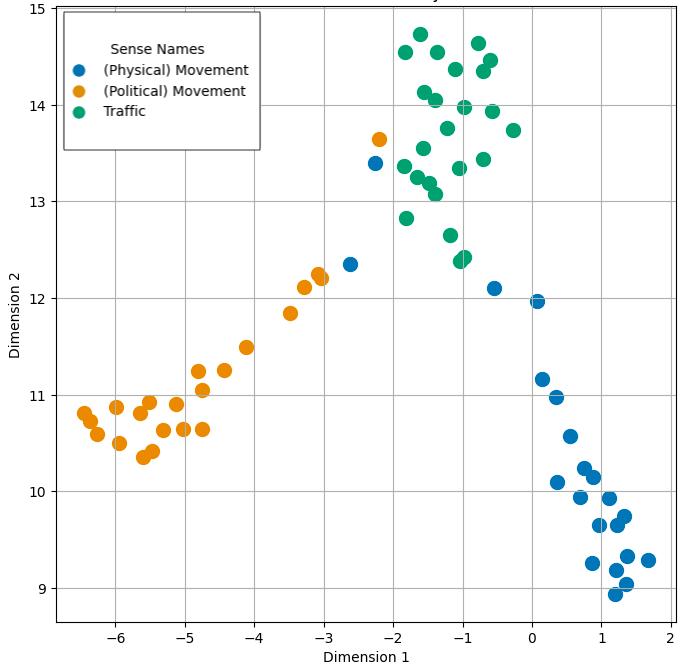}
    \caption{Word embeddings of the Polish word 'ruch' in sense-annotated sentences, visualized in 2D with UMAP. Interestingly, the resulting shape resembles a walking figure.}
    \label{fig:enter-label}
\end{figure}

\begin{onecolumn}

\section{Evaluating Annotation Efficiency}
\label{tab:pivc_more}
\begin{table*}[ht]
    \centering
    \renewcommand{\arraystretch}{1.2}
    \small
    \begin{tabular}{|*{8}{c|}}
        \hline
        \multirow{2}{*}{\textbf{Lang}} & \multirow{2}{*}{\textbf{Word}} & \multicolumn{3}{c|}{\textbf{Sense Definitions}}  
        & \multicolumn{3}{c|}{\textbf{Lift (\%)}} \\ 
        \cline{3-8}
        & & 1 & 2 & 3 & 1 & 2 & 3 \\ 
        \hline
        
        \multirow{5}{*}{\textbf{KN}} & {\includegraphics[height=0.85em]{latex/figs/kn_w1.png}} & Foot & Under & - & 269 & 141 & - \\ \cline{2-8}
        & {\includegraphics[height=0.85em]{latex/figs/kn_w2.png}} & Opinion & Religion & - & 104 & 900 & - \\ \cline{2-8}
        & {\includegraphics[height=0.85em]{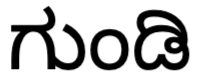}} & Pit/Hole & Bullet & - & 269 & 141 & - \\ \cline{2-8}
        & {\includegraphics[height=0.85em]{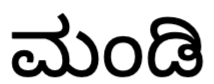}} & Market & Knee & - & 167 & 223 & - \\ \cline{2-8}
        & {\includegraphics[height=0.85em]{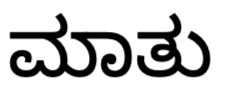}} & Word & Conversation & - & 289 & 128 & - \\ \hline
        
        \multirow{5}{*}{\textbf{MR}} & {\includegraphics[height=0.85em]{latex/figs/mr_w1.png}} & Juice & Interest & - & 228 & 288  & - \\ \cline{2-8}
        & {\includegraphics[height=0.85em]{latex/figs/mr_w2.png}} & Answer & North & - & 221 & 225 & - \\ \cline{2-8}
        & {\includegraphics[height=0.85em]{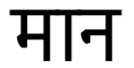}} & Respect & Approval & - & 218 & 235 & - \\ \cline{2-8}
        & {\includegraphics[height=0.85em]{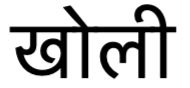}} & Room & Depth & - & 147 & 124 & - \\ \cline{2-8}
        & {\includegraphics[height=0.85em]{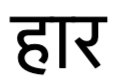}} & Necklace & Defeat & - & 106 & 149 & - \\
        \hline
        
        \multirow{5}{*}{\textbf{PA}} & {\includegraphics[height=0.85em]{latex/figs/pa_3.png}} & Bullet & Pill & -  & 107 & 1364 & - \\ 
        \cline{2-8}
        & {\includegraphics[height=0.85em]{latex/figs/pa_w2.png}} &  Thought & Intention & - & 235 & 884 & - \\ 
         \cline{2-8}
        & {\includegraphics[height=0.85em]{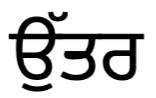}} & North & Response & Descend & 210 & 438 & 156 \\ \cline{2-8}
        & {\includegraphics[height=0.85em]{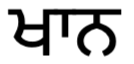}} & Khan (name) & Mine & - & 128 & 211 & - \\ \cline{2-8}
        & {\includegraphics[height=0.85em]{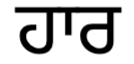}} & Defeat & Necklace & - & 129 & $\infty$ (prior = 0) & - \\  \hline
        
        \multirow{5}{*}{\textbf{UR}} & {\includegraphics[height=0.85em]{latex/figs/ur_w1.png}} &  Gold & Sleep & - & 161 & 232 & - \\ \cline{2-8}
        & {\includegraphics[height=0.85em]{latex/figs/ur_w2.png}} &  Thanks & Sugar & - & 106 & 1414 & - \\ \cline{2-8}
        & {\includegraphics[height=0.85em]{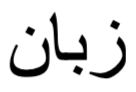}} & Language & Tongue & - & 119 & 358 & - \\ \cline{2-8}
        & {\includegraphics[height=0.85em]{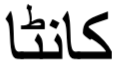}} & Thorn & Fork & - & 108 & 808 & - \\ \cline{2-8}
        & {\includegraphics[height=0.85em]{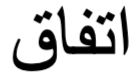}} & Opportunity & Agreement & Coincidence & 685 & 155 & 364 \\ 
        \hline
        
    \end{tabular}%
    
    \caption{Measured improvement over random chance (Lift) in semi-automated sentence selection over all evaluated words.}
    \label{tab:pivc_results}
\end{table*}

\section{WiC sentence pairing}


\begin{table*}[h]
    \small
    \centering
    \begin{tabular}{|l|llllllllll|}
    \hline
        Language & AZ & KN & KO & MR & PL & PA & SW & TE & UR & VI \\ \hline
        Sent Pairs (Train) & 20,409 & 20,298 & 5,703 & 19,368 & 13,516 & 26,237 & 7,312 & 23,115 & 14,018 & 5,153 \\ \hline
        Sent Pairs (Dev) & 5,649 & 5,627 & 1,018 & 5,175 & 3,562 & 7,025 & 2,165 & 5,861 & 3,450 & 751 \\ \hline
        Sent Pairs (Test) & 5,434 & 4,809 & 656 & 4,194 & 3,103 & 5,749 & 1,100 & 5,500 & 3,210 & 1,397 \\ \hline
        Words (Train) & 42 & 42 & 20 & 45 & 47 & 39 & 16 & 36 & 28 & 8 \\ \hline
        Words (Dev) & 22 & 22 & 11 & 24 & 25 & 21 & 9 & 19 & 15 & 5 \\ \hline
        Words (Test) & 22 & 21 & 9 & 22 & 24 & 19 & 7 & 18 & 14 & 4 \\ \hline
        Words in All Splits & 13 & 13 & 6 & 14 & 15 & 12 & 5 & 11 & 9 & 3 \\ \hline

    \end{tabular}
    \caption{Amounts of sentence pairs and unique polysemous target words in the train-dev-test splits of our constructed WiC datasets.}
    \label{tab:wic statistics}
\end{table*}

\end{onecolumn}

\end{document}